\documentclass[letterpaper]{article} 
\usepackage{aaai25}  
\usepackage{times}  
\usepackage{helvet}  
\usepackage{courier}  
\usepackage[hyphens]{url}  
\usepackage{graphicx} 
\usepackage{natbib}  
\usepackage{caption} 
\usepackage{upgreek}
\usepackage{amsmath}
\usepackage{bm}

\usepackage{algpseudocode}
\usepackage[ruled]{algorithm2e}

\usepackage[table,xcdraw]{xcolor}
\usepackage{colortbl}
\usepackage{pifont}
\usepackage[utf8]{inputenc} 
\usepackage[T1]{fontenc}    
\usepackage{url}            
\usepackage{booktabs}       
\usepackage{amsfonts}       
\usepackage{nicefrac}       
\usepackage{microtype}      
\usepackage{xcolor}         

\usepackage{graphicx}
\usepackage{floatrow}
\usepackage{subfig}
\usepackage{caption}
\usepackage{float}
\usepackage{newfloat}
\usepackage{listings}
\DeclareCaptionStyle{ruled}{labelfont=normalfont,labelsep=colon,strut=off} 
\floatstyle{ruled}
\newfloat{listing}{tb}{lst}{}
\floatname{listing}{Listing}
\title{Parallel-Learning of Invariant and Tempo-variant Attributes of Single-Lead Cardiac Signals: PLITA}
\author{
    Adrian Atienza,
    Jakob E. Bardram,
    Sadasivan Puthusserypady}
\affiliations{
    Technical University of Denmark\\

    \{adar, jakba, sapu\}@dtu.dk 
}

\usepackage{acronym}
\usepackage[acronym]{glossaries}
\newacro{SSL}{Self-Supervised Learning} 
\newacro{CHRONOS}{Contrasting Heads Represent Opposed Natures of Signals}
\newacro{ML}{Machine Learning}
\newacro{ECG}{electrocardiogram} 
\newacro{AFib}{Atrial Fibrillation}
\newacro{DEBS}{Distilled Encoding Beyond Similarities}
\newacro{EEG}{electroencephalogram}
\newacro{CPC}{Contrastive Predictive Coding}
\newacro{PCLR}{Patient Contrastive Learning}
\newacro{EMA}{exponential moving average}
\newacro{SBnCL}{Subject-Based non Contrastive Learning}
\newacro{SVC}{ Support Vector Classificatier}
\newacro{BYOL}{Boostrap Your Own Latent}
\newacro{DINO}{Self-Distillation with no Labels}
\newacro{PAR}{Pondered Average Representation}
\newacro{SHHS}{Sleep Heart Health Study}
\newacro{ViT}{Vision Transformer}
\newacro{MLP}{Multilayer Perceptron}
\newacro{MIT-ARR}{MIT-BIH Arrhythmia Database}
\newacro{MIT-AFIB}{MIT-BIH Atrial Fibrillation Database}
\newacro{PCA}{Principal Component Analysis}
\newacro{MAE}{Masked Autoencoders}
\newacro{BYOL}{Bootstrap Your Own Latent}
\newacro{ReLU}{rectified linear unit}
\newacro{PSG}{Polysomnography}
\newacro{SOTA}{state-of-the-art}
\newacro{TF-C}{Time-Frequency Consistency}
\newacro{Cinc2017}{Physionet Challenge 2017}
\newacro{NSRR}{National Sleep Research Resource}
\newacro{CLOCS}{Contrastive Learning of Cardiac Signals Across Space}
\newacro{DEAPS}{Distilled Embedding for Almost-Periodic Time Series}
\newacro{SOTA}{state-of-the-art}
\newacro{VIC-REG}{Variance-Invariance-Covariance Regularization}
\newacro{SGD}{Stochastic Gradient Descent}
\newacro{LOO}{Leave-One-Out}
\newacro{DIVA}{Disentangling Invariant and tempo-Variant Attributes}
\newacro{SIE}{Split Invariant-Equivariant}
\newacro{MIT-PSG}{MIT-BIH Polysomnographic Database}
\newacro{SHAP}{SHapley Additive exPlanations}
\newacro{GRU}{Gated Recurrent Unit}
\newacro{PLITA}{Parallel-Learning of Invariant and Tempo-variant Attributes}
\newacro{ISL}{Intra-inter Subject Self-Supervised Learning}
\newacro{ST-MEM}{Spatio-Temporal Masked Electrocardiogram Modeling}
\newacro{ASTCL}{Adversarial Spatiotemporal Contrastive Learning}

\begin{document}
\maketitle

\begin{abstract}
Wearable sensing devices, such as Holter monitors, will play a crucial role in the future of digital health.
Unsupervised learning frameworks such as \acf{SSL} are essential to map these single-lead \ac{ECG} signals with their anticipated clinical outcomes.
These signals are characterized by a tempo-variant component whose patterns evolve through the recording and an invariant component with patterns that remain unchanged.
However, existing \ac{SSL} methods only drive the model to encode the invariant attributes,
leading the model to neglect tempo-variant information which reflects subject-state changes through time.
In this paper, we present \acf{PLITA}, a novel \ac{SSL} method designed for capturing both invariant and tempo-variant \ac{ECG} attributes. 
The latter are captured by mandating closer representations in space for closer inputs on time.
We evaluate both the capability of the method to learn the attributes of these two distinct kinds, as well as \ac{PLITA}'s performance compared to existing SSL methods for \ac{ECG} analysis. \ac{PLITA} performs significantly better in the set-ups where tempo-variant attributes play a major role.
\end{abstract}

%

\section{Introduction}
The wearable sensing field has seen remarkable advancements in recent years. By enabling real-time data collection on physiological parameters, these devices will play a crucial role in the future of digital health. 
Among these wearable sensors, the Holter monitor captures the cardiac machinery as single-lead \ac{ECG} signals.
Leveraging the information that is accommodated in these signals has the potential of providing outstanding benefits: (i) Facilitating the early identification of irregular heart rhythms, such as \acf{AFib}, (ii) Simplifying the diagnostic process and minimizing the necessity for comprehensive testing \citep{s_ecg_1}, and (iii) Enabling users to engage proactively in tracking their heart health by offering instant access to health data and insights \citep{s_ecg_2}.
Generic models are mandated to map these data with their anticipated clinical outcomes. These models should compute informative single-lead \ac{ECG} representations applicable to several downstream tasks and be optimized using large volumes of unlabelled data. This makes \acf{SSL} framework particularly well-suited for addressing this clinical challenge.
\\

\begin{figure}[H]
\centering
{\fbox{\includegraphics[width=0.85\linewidth]{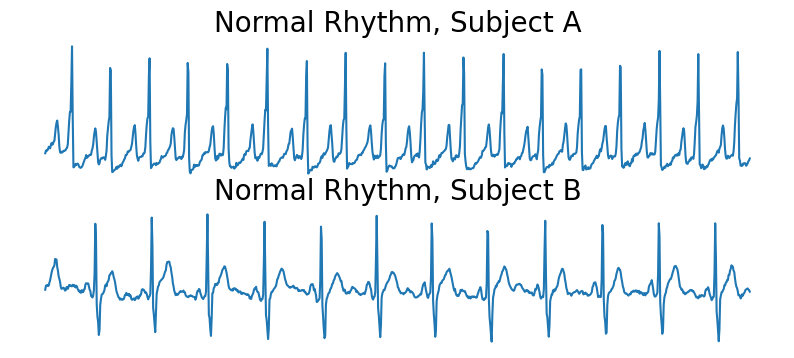}}}
\caption{A pair \ac{ECG} strips from distinct subjects are shown. The signal morphology accommodates a strong stationary component with visible differences between the subjects.} 
\label{fig:ecgs}
\end{figure}

\noindent 
To ensure blood reaches the entire body, each heart periodically repeats a sequence of actions, i.e., contractions, relaxations, and repolarizations, involving its different chambers with clockwise precision.
This unique execution of actions results in signals that exhibit a strong stationary component which is distinctive among hearts, as shown in Figure \ref{fig:ecgs}. These stationary attributes accommodate meaningful information such as the subject`s gender \citep{age-gender} or tendency to cardiac arrhythmia~\citep{baselines}.
In parallel, the functioning of the heart evolves, since it has to adapt to the individual's temporary needs or it may fall into arrhythmias. This evolution, also captured in the recordings, adds a non-stationary component to the data. 
Therefore, single-lead \ac{ECG} signals are characterized by two components of distinct kinds. The stationary component remains unchanging over time. Conversely, the non-stationary component is time-sensitive and its patterns evolve through the recording. This paper will refer to these components as invariant and tempo-variant attributes, respectively.
\\
%

\noindent Current \ac{SSL} techniques designed for single-lead \ac{ECG} processing~\citep{PCLR, clocs, mix_up} enforce the model to understand and encode the signal invariant patterns following a Contrastive Learning \citep{simclr} approach.
Despite each having its uniqueness, they all consider non-overlapping signal strips from the same subject as positive pairs and enforce similar representations between them.
These studies demonstrate that by exploiting the non-stationarity nature of the data, the invariant information of the Single-Lead \ac{ECG} signals are better captured within the representations 
rather than creating two versions of the same input using data augmentation techniques.
This common strategy is aligned with other time series \ac{SSL} works, \citep{video1, video2, video3}, where non-overlapping frames from the same video are considered as positive pairs during the training procedure.\\
%

\noindent However, solely focusing on driving the model to capture the invariant attributes only covers a part of the whole picture.
In other words, simply mandating similar representations from inputs belonging to the same recording will lead to the model neglecting the tempo-variant attributes, and thereby these changes over time. It leads to a loss of meaningful information 
that is particularly valuable in specific scenarios aimed at identifying occasional cardiovascular events that occur at irregular intervals throughout the recording, such as detecting \ac{AFib} or classifying sleep stages.\\

\begin{figure}[t]
\vskip 0.2in
\begin{center}
\centerline{\fbox{\includegraphics[width=0.5\columnwidth]{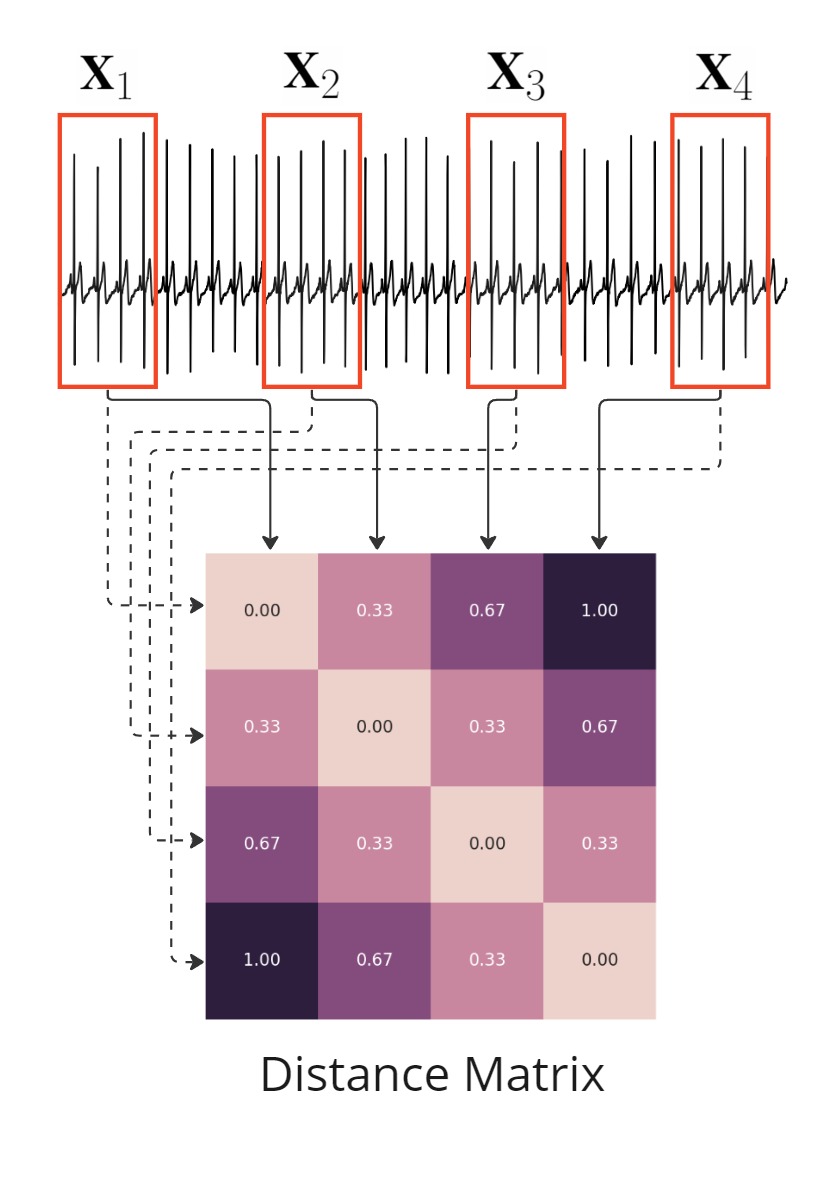}}}
    \caption{Considering time-sorted inputs equally spaced in time ($X_1 \dots X_4$), representations of nearby inputs in time are expected to be closer than time-distant ones.}
    \label{fig:distances}
\label{fig:distances}
\end{center}
\vskip -0.2in
\end{figure}

\noindent To address this drawback, this paper presents \acf{PLITA}, a novel \ac{SSL} method designed to represent both the invariant and the tempo-variant attributes of single-lead \ac{ECG} signals.
%
The proposed \ac{PLITA} approach is consistent with methodologies like \ac{SIE}~\citep{sie}, where the model is designed to capture attributes of two different kinds.
\ac{PLITA} compels its learning model to integrate tempo-variant attributes by ensuring that the representations adhere to a coherent principle: representations of temporally proximate inputs should be closer than those of temporally distant inputs. This concept is depicted in Figure~\ref{fig:distances} and is encapsulated in the innovative ``Tempo-variant Loss Function'' ($\mathcal{L}_{tv}$), which forms an integral part of the training objective.\\

\noindent In this study, we hypothesize that: 
(i) Tempo-variant attributes contain significant information that is distinct from the information conveyed by invariant ones,
(ii) Simply using the tempo-variant attributes as a source of natural variance limits the potential of the representation in several downstream tasks,
(iii) These attributes can also be incorporated within the representations by the proposed $\mathcal{L}_{tv}$ loss function, and  
(iv) By encoding these attributes, the model performance improves significantly in set-ups where the tempo-variant attributes play an important role. To assess these hypotheses, we have conducted three experiments that require the invariant or/and the tempo-variant attributes to be encoded within the representations: 
\begin{enumerate}
    \item \textbf{\ac{AFib} Classification.} \ac{AFib} episodes can be sporadic in time. Moreover, the susceptibility of an individual to this disease is indicated in the baseline signal~\cite{baselines}.
    Therefore, both the invariant and tempo-variant attributes will play a role in this task. 
    \item \textbf{Sleep Stages Classification.} Various sleep stages occur throughout the sleep cycle, regardless of the individual. Consequently, it is essential for the model to capture the tempo-variant attributes to successfully perform this task.
    \item \textbf{Gender Identification task.} Gender-related information will be persistent within the data throughout the recording. Therefore, driving the model to encode the invariant attributes will be required in this task. \\
\end{enumerate}

\noindent We have evaluated the performance of \ac{PLITA} against the \ac{SOTA} methods designed for sinle-lead \ac{ECG} processing. The findings indicate a marked enhancement in performance on downstream tasks where tempo-variant characteristics are influential. Furthermore, the model demonstrates robust results in gender classification, confirming also the presence of invariant features within the representations. Additionally, we assess whether both tempo-variant and invariant features are reliably captured in the representations and if they fulfill their intended function in downstream tasks. \\

\noindent In summary, the contributions of this paper are: 
\begin{itemize}
\item We demonstrate through empirical results that the tempo-variant attributes of cardiac signals are not merely a source of natural variation for efficient invariant attribute learning, as assessed in previous studies, but must also be integrated into the representations.

\item We introduce \ac{PLITA}, a novel \ac{SSL} method that takes apart from existing \ac{ECG} processing \ac{SSL} methods by driving the model to encode the tempo-variant attributes via the novel $\mathcal{L}_{tv}$ function.

\item We establish a new approach for harnessing tempo-variant features. We hypothesize that this may motivate future \ac{SSL} work not only applied to \ac{ECG} analysis but also broadly to other time series data.
\end{itemize}

\begin{figure*}[t]
\vskip 0.2in
\begin{center}
\centerline{\fbox{\includegraphics[width=0.6\linewidth]{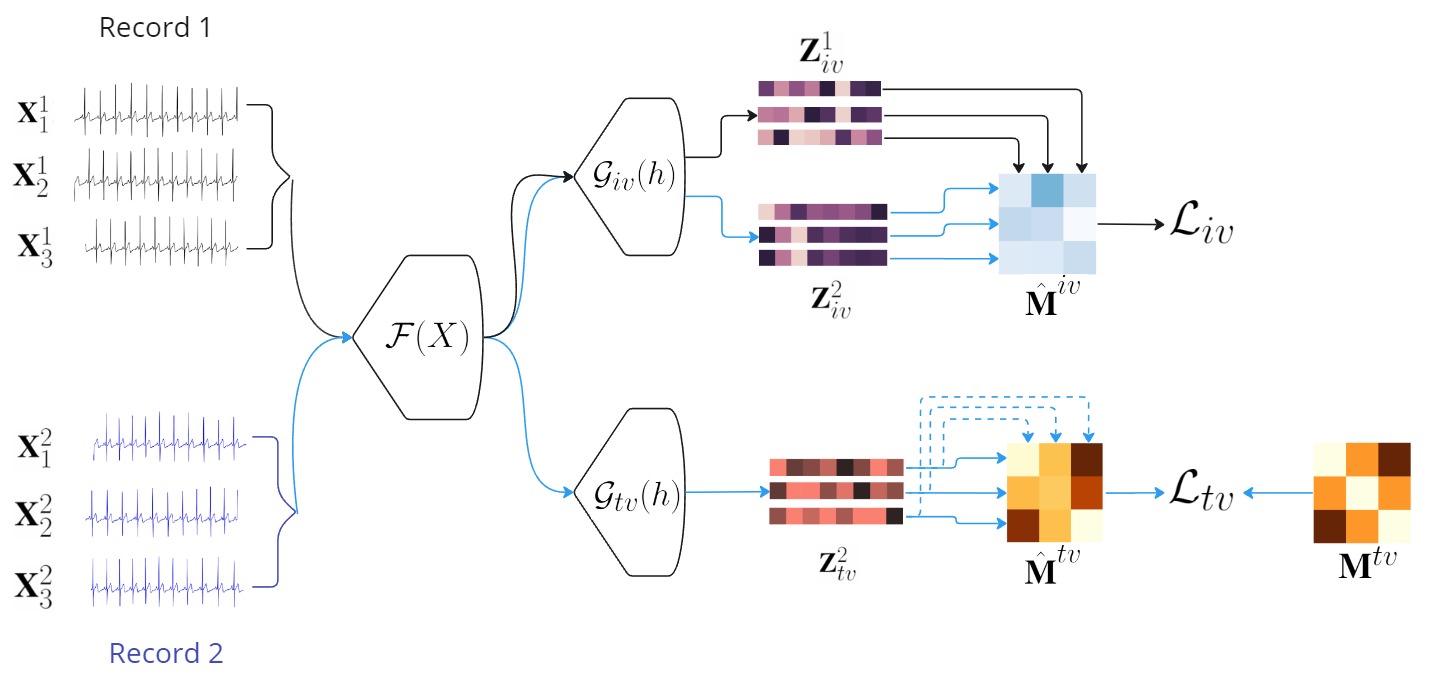}}}
\caption{PLITA illustrated. Built on top of BYOL, PLITA includes both a student and a teacher network. For the sake of clarity, the teacher network is not included in the illustration. The losses are computed between representation triplets from both networks that process data equally. While $\mathcal{L}_{iv}$ is computed between a set of $N$ time series representations belonging to different records (displayed in black and blue colors), $\mathcal{L}_{tv}$ is computed between representations belonging to the same record. All inputs belong to the same subject. The encoder ($\mathcal{F}(X)$) is saved at the end of the training procedure and used for downstream tasks.} 
\label{fig:PLITA}
\end{center}
\vskip -0.2in
\end{figure*}

\section{Related Work}

\subsection{Tempo-Variant attributes in Time Series}
Time series data represents the evolution of an object of interest over time. Existing methods  \citep{video1, video2, video3} leverage naturally occurring variations in consecutive frames to define positive pairs, avoiding heavy reliance on data augmentation techniques. 
They provide evidence that this approach enhances the model’s effectiveness in handling downstream tasks. Another example of how the availability of a sequence of inputs can improve model training is Siamese Masked Autoencoders work \cite{siam_auto}, which extends Masked Autoencoders \cite{masked} by optimizing the model to reconstruct the subsequent frames instead of the actual one.

\subsection{SSL in Single-Lead \ac{ECG} Signal Processing}
Existing \ac{SSL} methods tailored for single-lead \ac{ECG} data are aligned with \ac{SSL} methods for video processing. 
They leverage the data shifts across time for enhancing the learning of the invariant attributes.
They all utilize the Contrastive Learning \citep{simclr} as a common framework, considering non-overlapping inputs as positive pairs.
The details of the positive-pair selection strategy set these methods apart:
(i) The Mixing-Up method~\cite{mix_up} introduces a more tailored data augmentation product of two time series from the same recording. 
(ii) \ac{CLOCS} \cite{clocs} utilizes two consecutive \ac{ECG} time strips as positive pairs, and
(iii) \ac{PCLR} \cite{PCLR} which considers two time strips from the same subject but different recordings.
\ac{PLITA} inherits the \ac{PCLR} strategy by defining positive pairs from distinct recordings to ensure the representation of invariant attributes, as it outperforms the other methods (See Evaluation).\\

\noindent While the previous studies only focus on the invariant attributes, \ac{ISL} \cite{intra_inter} mandates representing alterations between consecutive beats.
\ac{PLITA} contemplates the tempo-variant information spotted among temporally sparse inputs, not just between consecutive beats. The proposed method incorporates a novel loss function that drives the model to encode this information by comparing these delayed inputs.

\subsection{SSL in 12-Lead \ac{ECG} Signal Processing}
The most recent work on the \ac{ECG} field focuses on 12-Lead signals. Having multiple leads opens up the spatial dimension and thus the range of possibilities when designing methods to process this kind of data \cite{stmem, 12-lead2}.
However, it is not possible to adapt them to single-lead \ac{ECG} processing. Part of their potential is based precisely on exploiting the spatial dimension, which is not available in the wearable sensing field in which this study is placed. Therefore, they are not included as baselines during the evaluation of \ac{PLITA}. 
\\

\section{\acf{PLITA}}

The aim of \ac{PLITA} is to simultaneously drive the model to recognize both the invariant and tempo-variant attributes and encode them in the representations. Its workflow is illustrated in Figure~\ref{fig:PLITA}. 
Here, \textit{N} inputs equally delayed in time within a window size \textit{W} are sampled from two records belonging to the same subject.
The inputs are given to the learning model, displayed as $\mathcal{F}(X)$.
The model computes the representations (denoted as \textit{h}),
which are passed through the invariant and tempo-variant projectors (denoted as $\mathcal{G}_{iv}$ and $\mathcal{G}_{tv}$, respectively).
This workflow is mimicked by the teacher network.
The invariant distance matrix, denoted as  $\hat{\textbf{M}}^{iv}$, is calculated between the invariant projections of both recordings ($\textbf{z}^1_{iv}$ and $\textbf{z}^2_{iv}$,).
The invariant loss function, $\mathcal{L}_{iv}$, minimizes the values of $\hat{\textbf{M}}^{iv}$.
Parallel to this, the tempo-variant distance matrix, $\hat{\textbf{M}}^{tv}$, is calculated between the tempo-variant projections from the same recordings ($\textbf{z}^2_{tv}$). 
This matrix is compared with the ideal tempo-variant distance matrix ($\textbf{M}^{tv}$) by the tempo-variant loss function, $\mathcal{L}_{tv}$.
\noindent The student encoder ($\mathcal{F}(X)$) is the module used when addressing the downstream tasks. The other components are discarded after the training.

\subsection{Projecting the Invariant and Tempo-variant into Distinct Spaces}\label{sec:two_projectors}

\ac{PLITA} is matched with other existing methods such as \ac{SIE}, which are designed to integrate various types of attributes into a singular representation.
%
Analogously, \ac{PLITA} incorporates two projectors ($\mathcal{G}_{iv}$ and $\mathcal{G}_{tv}$) to
project the representations into two different spaces and avoid some conflicting goals during the training procedure. 
These arise from the fact that while $\mathcal{L}_{iv}$ minimizes the distance between the representations, $\mathcal{L}_{tv}$ encourages these distances between time-sorted inputs to occur and follow a spatial order. \\

\noindent The equivariant attributes defined in \ac{SIE}'s method are not related to the tempo-variant ones, nor how they are considered in each respective method. \ac{PLITA} also differs from \ac{SIE} in not splitting the representations nor incorporating the $\mathcal{L}_{reg}$ term. Variations in baseline \ac{ECG}s are crucial for identifying spontaneous episodes of cardiovascular diseases \citep{baselines}. We contend that a holistic consideration of the representation will yield more precise tempo-variant projections. Moreover, the divergent objectives of $\mathcal{L}_{iv}$ and $\mathcal{L}_{tv}$ are anticipated to be adequate in preventing mode collapse between the two projections.
Yet the implications of these two decisions have been explored (Refer to Appendix).

\subsection{Capturing Invariant and Tempo-variant Attributes}

This section provides a detailed description of the main technical contribution of this work, which is \ac{PLITA}'s proposed parallel learning for driving the model to capture both the invariant and the tempo-variant attributes of cardiac signals. 
\subsubsection{Capturing Invariant Attributes:} Similarly to \ac{PCLR} method, PLITA drives the model to encode invariant attributes by employing inputs from different recordings from the same subject. Nevertheless, \ac{PLITA} approaches this objective in a Non-Contrastive Learning fashion, being built on top of \ac{BYOL} framework. Both decisions are supported by superior performance in the invariant downstream task of gender classification (see Evaluation).
We use the ``cosine similarity'' for calculating the distance matrix between the invariant projections, denoted as $\hat{\textbf{M}}^{iv}$ and it defined as the following: 

\begin{equation}
\hat{\textbf{M}}^{iv}_{i, j} = 1 - \frac{(\bm{\upzeta}_{iv}^1)^i \cdot \mathcal{Q}_{iv}((\textbf{z}_{iv}^2)^j)} {\max \left(\left\|(\bm{\upzeta}_{iv}^1)^i\right\|_2 \cdot\left\|\mathcal{Q}_{iv}((\textbf{z}_{iv}^2)^j)\right\|_2, \epsilon\right)} \text {,}
\end{equation}

where $(\upzeta_{iv}^1)^i$ and $\mathcal{Q}_{iv}((\textbf{z}_{iv}^2)^j)$ are the invariant outputs of the teacher and student networks respectively, for the inputs with index $i$ and $j$ drawn from the records 1 and 2. Note that the \ac{BYOL} framework features both a teacher and a student network. These parallel networks and the student projector are not illustrated in Figure~\ref{fig:PLITA} for the sake of clarity. 
Yet in practice, we calculate each loss function by comparing the output of the student prediction, ($\mathcal{Q}({\textbf{z}}$)) with the subsequent output of the teacher projector ($\bm{\upzeta}$). This logic also applies to the $\mathcal{L}_{iv}$ introduced below.

\noindent Making the invariant projections similar implies minimizing the values of $\hat{\textbf{M}}^{iv}_{i, j}$, therefore we define the $\mathcal{L}_{iv}$ as;
\begin{equation}
\mathcal{L}_{iv} = \frac{1}{N^2} \sum_{i}^N \sum_{j}^N \hat{\textbf{M}}^{iv}_{i, j}.
\end{equation}

\begin{table*}[t]
\centering
\resizebox{\textwidth}{!}{%
\begin{tabular}{lcccccccccccc}
\toprule
\rowcolor{gray!20}
\textbf{Evaluation Task} & 
\multicolumn{4}{c}{{\textbf{AFIB Classification}}} & 
\multicolumn{4}{c}{{\textbf{Sleep Stage Classification}}} & 
\multicolumn{4}{c}{{\textbf{Gender Identification}}} \\
 
\cmidrule(lr){2-5} \cmidrule(lr){6-9} \cmidrule(lr){10-13}
\textbf{Pre-Train Dataset} & 
\multicolumn{2}{c}{\textbf{SHHS}} & 
\multicolumn{2}{c}{\textbf{Icentia}} & 
\multicolumn{2}{c}{\textbf{SHHS}} & 
\multicolumn{2}{c}{\textbf{Icentia}} & 
\multicolumn{2}{c}{\textbf{SHHS}} & 
\multicolumn{2}{c}{\textbf{Icentia}} \\
 
\cmidrule(lr){2-3} \cmidrule(lr){4-5}
\cmidrule(lr){6-7} \cmidrule(lr){8-9}
\cmidrule(lr){10-11} \cmidrule(lr){12-13}
\textbf{Method / Metric} & 
\textit{Accu. (\%)} & \textit{F1 Score} & \textit{Accu. (\%)} & \textit{F1 Score} & 
\textit{Accu. (\%)} & \textit{AUC} & \textit{Accu. (\%)} & \textit{AUC} & 
\textit{Accu. (\%)} & \textit{AUC} & \textit{Accu. (\%)} & \textit{AUC} \\
 
\midrule
\rowcolor{gray!10}
\textbf{PCLR} & 76.4 & 73.7 & 73.7 & 73.6 & 
71.6 & 0.75 & 72.3 & 0.77 
& 76.4 & \textbf{0.84} & 66.5 & \textbf{0.74} \\

\textbf{Mix-Up} & 73.4 & 72.3 & 62.9 & 57.3 
& 73.8 & 0.79 & 72.6 & 0.75 
& 70.4 & 0.76 & 64.2 & 0.69\\

\rowcolor{gray!10}
\textbf{CLOCS} & 75.7 & 73.8 & 73.6 & 72.7 
& 73.2 & 0.78 & 72.0 & 0.74 
& 70.4 & 0.76 & 65.3 & 0.7 \\

\textbf{BYOL} & 76.6 & 74.8 & 75.3 & 72.5 
& 72.9 & 0.77 & 73.3 & 0.78  
& \textbf{76.7} & 0.83 & \textbf{66.7} & 0.72 \\

\rowcolor{gray!10}
\textbf{Ti-MAE} & 72.1 & 70.9 & 52.9 & 60.0
& 69.3 & 0.61 & 69.3 & 0.66 
& 60.0 & 0.6 & 60.0 & 0.61\\

\textbf{Siam Auto} & 76.5 & 73.0 & 53.3 & 70.1 
& 72.9 & 0.74 & 69.3 & 0.68 
& 73.3 & 0.8 & 55.4 & 0.47 \\
\rowcolor{gray!10}

\textbf{PLITA} & \textbf{80.7} & \textbf{78.4} & \textbf{80.0} & \textbf{78.2} 
& \textbf{75.3} & \textbf{0.81} & \textbf{74.8} & \textbf{0.8} 
& 76.5 & 0.83 & 66.5 & 0.72 \\
 
\bottomrule
\end{tabular}}
\vspace{5pt}
\caption{Evaluation Results for the three downstream tasks, using the pretrained model trained from both Icentia and SHHS datasets. The bold type indicates the best-performing method for each metric.}
\label{tab:eval}
\end{table*}

\subsubsection{Capturing Tempo-variant Attributes:}
\ac{PLITA} requires the model to compute representations in a spatial order aligned with the chronological sequence of time. In other words, the closer the inputs are on time, the closer the representations should be in space.
This desired behavior is modeled by the so-called ``Ideal Tempo-variant Distances Matrix'' ($\textbf{M}^{tv}$), which is defined as;
\begin{equation}
\textbf{M}_{i, j}^{tv}= \frac{|i - j|}{N - 1}, 
\end{equation}
where $i$ and $j$ are the indices of the inputs sorted in time, and \textit{N} is the number of inputs considered in each window size. Although we consider the ``cosine similarity'' as the distance metric between two representations from the same recording, 
other choices for distance metrics have been evaluated (Refer to Appendix).
These pair-wise distances are captured by $\hat{\textbf{M}}^{tv}$, which is computed as;
\begin{equation}
\hat{\textbf{M}}^{tv}_{i, j} = 1 - \frac{(\upzeta_{tv}^2)^i \cdot \mathcal{Q}_{tv}((\textbf{z}_{tv}^2)^j)} {\max \left(\left\|(\upzeta_{tv}^2)^i\right\|_2 \cdot\left\|\mathcal{Q}_{tv}((\textbf{z}_{tv}^2)^j)\right\|_2, \epsilon\right)} \text {, }
\end{equation}
where $(\bm{\upzeta}_{tv}^2)^i$  and $\mathcal{Q}_{tv}((\textbf{z}_{iv}^2)^j)$ are the tempo-variant outputs of the teacher and student networks respectively, for the inputs with index $i$ and $j$ drawn from the same record. $\hat{\mathbf{M}}^{tv}$ is scaled before calculating the $\mathcal{L}_{iv}$ as the following;
\begin{equation}
 \hat{\textbf{M}}^{tv}_{\prime} = a + \frac{(\hat{\textbf{M}}^{tv} -\min (\hat{\textbf{M}}^{tv}))(b - a)}{\max (\hat{\textbf{M}}^{tv})-\min (\hat{\textbf{M}}^{tv})},
\end{equation}\\
where $a = 1/(N - 1)$ and $b = 1$. By scaling $\hat{\textbf{M}}^{tv}$, we do not only ensure that the values of $\hat{\textbf{M}}_\prime^{tv}$ and $\textbf{M}_{tv}$ lie within the same range, 
but also we alleviate the constraints imposed by $\mathcal{L}_{tv}$. 
\ac{PLITA} just mandates the representations to follow a tempo-spatial order
without imposing a constant distance for every set of inputs. 
This constant distance would be a problem since the same magnitude of variance can not be expected for each set in the batch.
The final $\mathcal{L}_{tv}$ that enforces the tempo-variant attributes to be represented into the representations is defined as;
\begin{equation}
\mathcal{L}_{tv}=\frac{1}{N (N- 1)} \sum_{i}^N \sum_{j \neq i}^N\left((\mathbf{M}^{tv})_{i, j}-\hat{\textbf{M}}^{tv}_{\prime})_{i, j}\right)^2.
\end{equation}
Note that \ac{PLITA} does not take into account the diagonal terms, since the invariant features are expected to be modeled by the $\mathcal{L}_{iv}$ loss function. The evaluation of the tempo-variant loss term’s integration is presented in (See Ablation).

\section{Implementation Details}

To ensure the replication of the method, we meticulously outline the hyperparameter settings and the model architecture.\\

\noindent \textbf{Model Architecture:} The \ac{ViT}~\cite{vit} model is used for processing the single-lead \ac{ECG} signals. The input consists of a one dimensional 10-second signal sampled at 100 Hz. The patch size is set to 20. The model counts with 6 regular transformer blocks with 4 heads each and a dimension of 128.\\ 

\noindent \textbf{PLITA Implementation and Optimization}: The window size \textit{W} is set to 10 seconds. \textit{N} is set to 4, so 4 inputs are drawn from each window. The effect of both \textit{W} and \textit{N} is discussed in Section~\ref{sec:abla}.
Although we do not incorporate any data augmentation, the effect of it is discussed in the Appendix.
The projectors and predictors are implemented as a two-layer \ac{MLP} with a dimensionality of 512 and 256, respectively. 
The \ac{EMA} updating factor ($\tau$) is set to 0.995. 
The training procedure consists of 35,000 iterations. We use a batch size of 256, Adam~\cite{adam} with a learning rate of $3e-4$,  and a weight decay of $1.5e-6$ as the optimizer. The training procedure and the evaluations are performed on a desktop computer, with a Nvidia GeForce RTX 3070 GPU.

\section{Evaluation}\label{sec:eval}

In this section, we evaluate the performance of \ac{PLITA} compared with the most relevant existing \ac{SSL} methods for single-lead \ac{ECG} processing. We also have conducted a study on representations to assess whether the model’s learned representations effectively separate the invariant and tempo-variant attributes. Overall, the evaluation involves four distinct databases: \ac{MIT-AFIB} \cite{mit-afib}, \ac{MIT-PSG}, \cite{mit-psg}, \ac{Cinc2017} \cite{cinc2017} and \ac{SHHS} \cite{shhs1}. All databases are publicly available in Physionet \cite{physionet} and \ac{NSRR}.

\subsection{Comparison against \acf{SOTA}}\label{sec:sota}

The performance of the proposed \ac{PLITA} method has been compared against the three most relevant energy-based \ac{SOTA} methods, namely, (i) \ac{PCLR}~\cite{PCLR}, (ii) \ac{CLOCS} \cite{clocs}, and (iii) Mixing-Up \cite{mix_up}. 
Reconstruction methods such as (iv) Ti-MAE \citep{timae} or (v) Siamese Masked Autoencoderss \citep{siam_auto} (For more details about this latter implementation, refer to Appendix).
Finally, We have also included (vi) the \ac{BYOL} method tailored by \ac{ECG} processing by following the PCLR strategy for selecting the positive pairs.
To guarantee an equitable assessment, we have optimized the identical model employed in this study, maintaining consistent settings such as the optimizer, data, batch size, and iteration count. Each method has been trained in two distinct datasets, \ac{SHHS} \cite{shhs1,shhs2} and Icentia \citep{icentia}, that are composed of long-term single-lead \ac{ECG} recordings.

\paragraph{\ac{AFib} Classification:}
To assess the ability of the method to generalize different classes within the same record, given a limited number of labelled records, we have conducted a \ac{LOO} cross-validation across the 23 \ac{MIT-AFIB} subjects. This ensures no subject-overlapping between the training and validation sets which would significantly simplify AFib identification. A \ac{SVC} \cite{svc} is fitted on top of the representations. Table \ref{tab:eval} reflects the results, where it can be seen that \ac{PLITA} significantly outperforms the other methods.

\begin{figure*}[t]
\centering
\subfloat[Disentangling Study for invariant features in MIT-AFIB dataset.]
{\fbox{\includegraphics[width=0.485\linewidth]{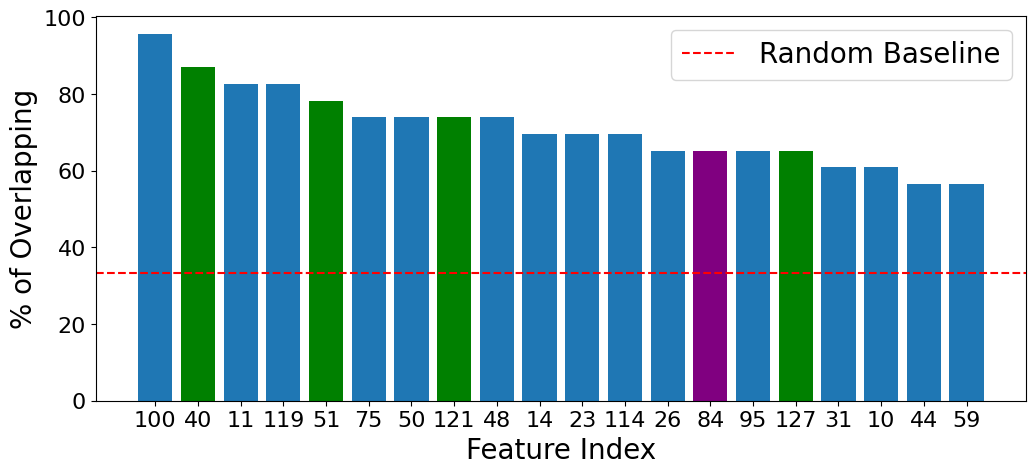}\label{fig:static_overlap}}}
\hspace{0.01\linewidth}
\subfloat[SHAP Analysis in gender classification.]
{\fbox{\includegraphics[width=0.302\linewidth]{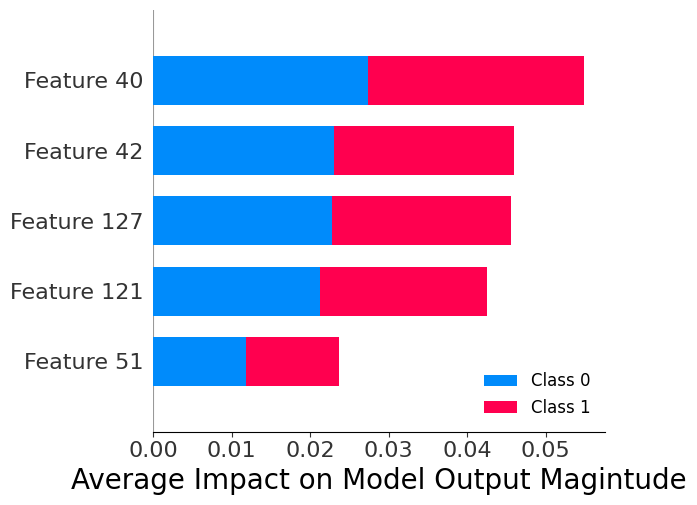}\label{fig:static_importance}}}
\caption{The features that play an important role in the gender classification task (displayed in Figure \ref{fig:static_importance}), are highlighted in green in Figure \ref{fig:static_overlap}. The feature that accounts for the AFib classification task (Figure \ref{fig:dynamic_importance}) is displayed in purple.}
\label{fig:shap2}
\end{figure*}

\begin{figure*}[t]
\centering
\subfloat[Disentangling Study for tempo-variant features in MIT-AFIB.]
{\fbox{\includegraphics[width=0.485\linewidth]{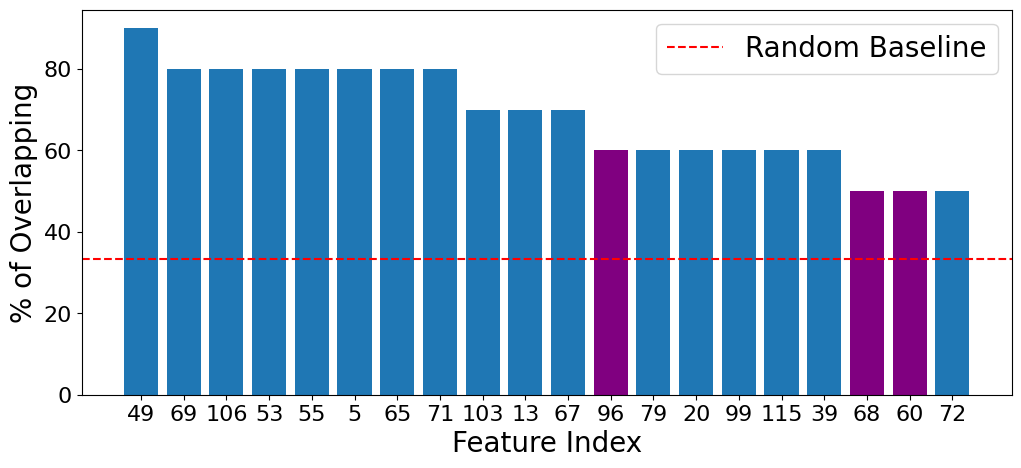}\label{fig:dynamic_overlap}}}
\hspace{0.01\linewidth}
\subfloat[SHAP Analysis in AFib classification.]
{\fbox{\includegraphics[width=0.302 \linewidth]{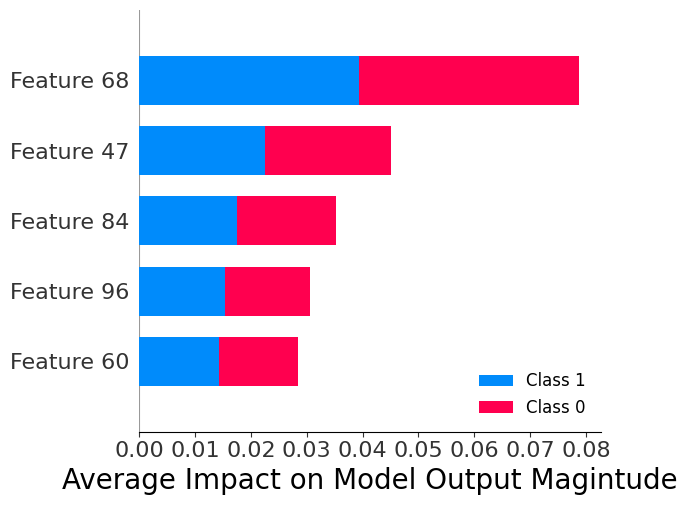}\label{fig:dynamic_importance}}}
\caption{The informative features in the AFib classification are displayed in Figure \ref{fig:dynamic_importance} and highlighted in purple in Figure \ref{fig:dynamic_overlap}.}
\label{fig:shap1}
\end{figure*}

\paragraph{Sleep Stage Detection:}
We have used the \ac{MIT-PSG} database in order to assess the capability of the representations to discriminate between Sleep and Wake classes.
Since the golden standard classification is performed every 30 seconds and our model has been optimized for processing 10-second signals, a \ac{GRU} \cite{gru} layer is fitted on top of the representations during 5 epochs for processing 30 seconds of data sequentially, in non-overlapping 10-second chunks. The pre-trained model is kept frozen. 
We have carried out a \ac{LOO} cross-evaluation for the 18 records contained in the dataset.
The outcomes of this analysis are detailed in Table \ref{tab:eval}. It is evident that \ac{PLITA} achieves a notably higher level of performance.

\paragraph{Gender Classification:}

We conducted a five-fold cross-validation over 1500 randomly-selected inputs from distinct subjects from the \ac{SHHS} database. A \ac{SVC} is fitted on top of the representations. Table \ref{tab:eval} shows that despite not achieving the best performance, \ac{PLITA} reaches competitive results.\\

\noindent Results conclude that training with Icentia data tends to yield worse results, possibly due to increased noise in the data. 
This decline in performance is more pronounced for reconstruction-based methods. 
\ac{PLITA} performs best in tasks that involve encoding tempo-variant attributes. 
The comparable results achieved in gender classification were also expected due to the only influence of the invariant attributes in this task and the identical manner of \ac{BYOL}, \ac{PCLR} and \ac{PLITA} to drive the model to encode them.

\subsection{Representation Study} \label{sec_shap}
The purpose of this two-phase experimental analysis is to verify that: (i) \ac{PLITA} prompts the model to encapsulate both invariant and tempo-variant attributes into a suite of discrete features that are consistent across different recordings, and (ii) The collection of invariant features is crucial for the gender classification task, as anticipated. Additionally, the tempo-variant attributes are highly relevant for \ac{AFib} classification (for more details on the \ac{AFib} experiment, see Appendix). 
%

\begin{figure*}[t]
\centering
\subfloat[AFib Classification.]
{\fbox{\includegraphics[width=0.37\linewidth]{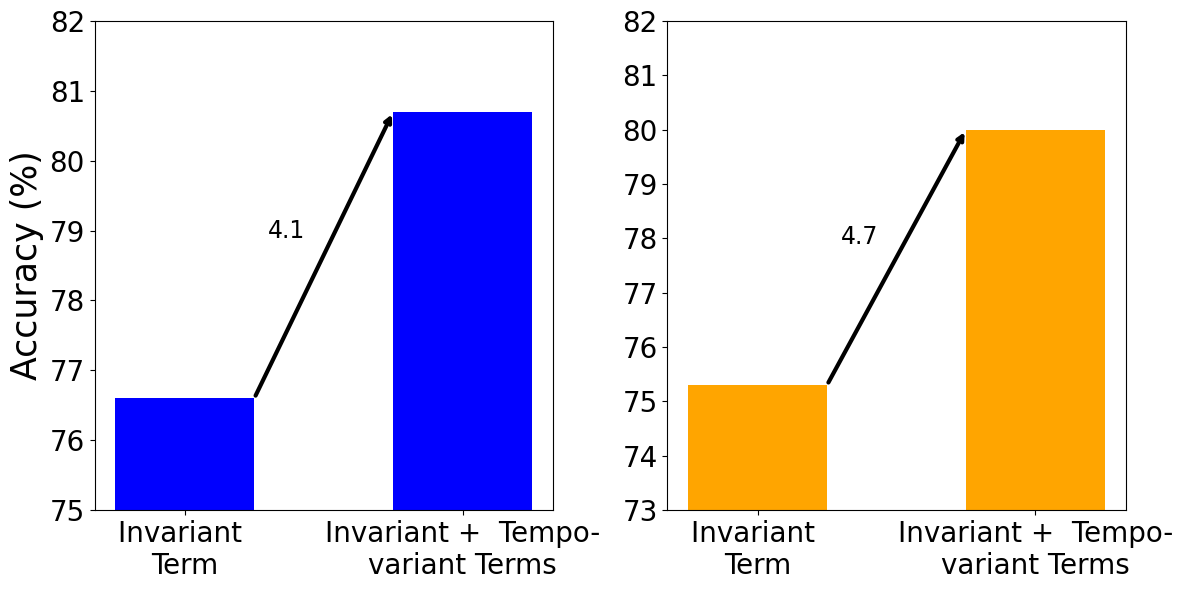}\label{fig:abla1}}}
\subfloat[Sleep Stage Classification]
{\fbox{\includegraphics[width=0.37\linewidth]{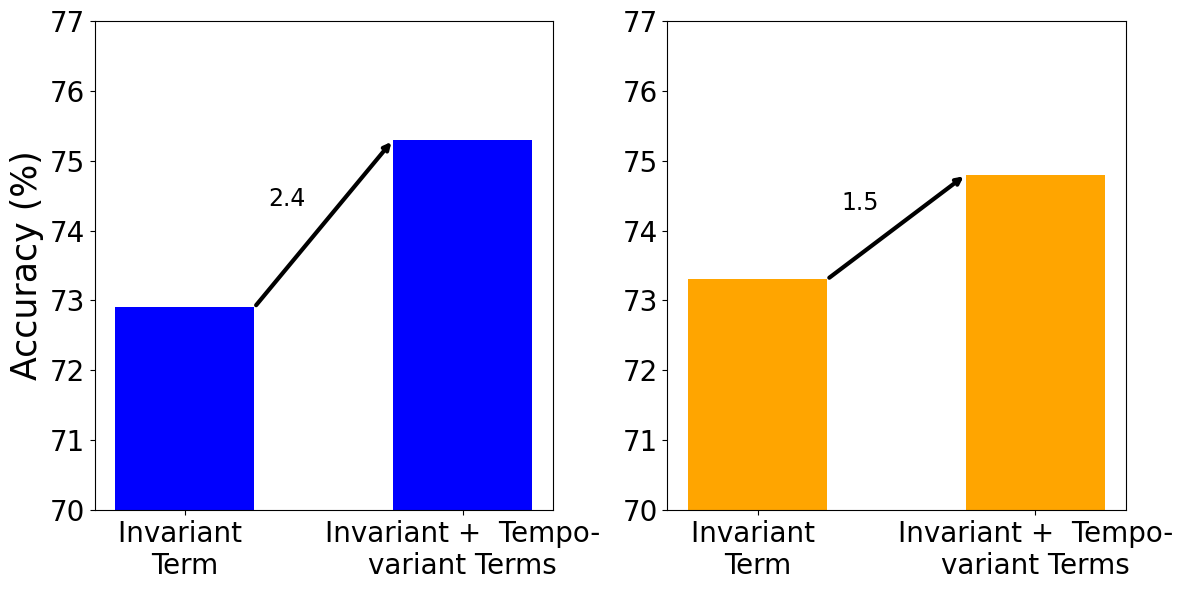}\label{fig:abla2}}}
\caption{Effect of incorporating the tempo-variant attributes in distinct tasks. Blue and Orange bars represent that the model has been pre-trained on SHHS and Icentia respectively.}
\label{fig:effect}
\end{figure*}

\paragraph{Disentangling Study:}
In order to verify that the two non overlapping sets of features that express the invariant and tempo-variant attributes across the different records are consistent, 
%
%
we have computed the representations of the 23 recordings from the \ac{MIT-AFIB} dataset.
Each feature value is normalized to ensure uniform value ranges.
Since it can be assumed that invariant and tempo-variant features will have low and high intra-record variance respectively, this variance is used as a measure of discretization. 
We cluster the 33\% of the features with the least variance as the invariant features while the 33\% of them with the highest variance are clustered as the tempo-variant ones.
Finally, we tallied the occurrences of each feature in both clusters for each recording. \\

\noindent Figure \ref{fig:static_overlap} shows the 20 features that appear most often as invariant features for each recording. The ratio of appearances goes from 95.7\% to 56.2\% so we can give it a statistical value since in a random baseline, the number of appearances would be around 33.3\%. This serves as evidence that the features which represent the invariant attributes of the \ac{ECG} signals are consistent across recordings. In a similar manner, Figure \ref{fig:dynamic_overlap} represents the 20 tempo-variant features, with a ratio of appearances from 90\% to 50\%. Therefore, it is assessed that the tempo-variant features are also consistent.

\paragraph{\acf{SHAP} Analysis:}

We have conducted a \ac{SHAP} analysis \cite{shap} for the \ac{AFib} and gender classification tasks.
Since the Disentangling study has been carried out in the \ac{MIT-AFIB} database, we used the \ac{AFib} and SR instances from \ac{Cinc2017}, adhering to the training and testing set proposed on it. For the gender task, we used the same dataset used in the evaluation. 
Figure \ref{fig:static_importance} and Figure \ref{fig:dynamic_importance} reflect how four and three of the five most important features are included among the invariant and tempo-variant sets, respectively. 
This result is of statistical relevance, since in a random baseline only $6\mathrm{e}{-3}$ features would be included in each of the 20-size set of features.\\

\noindent Notably, the third most important feature for AFib detection exhibits an invariant nature (Feature 84). Although at first sight, this may seem contradictory, 
this aligns \ac{PLITA} with the findings of other  studies \cite{baselines} which claim that invariant attributes present in \ac{ECG} baselines enable discretizing the subjects that are susceptible to suffer episodes.
%
%
%

\subsection{Discussion of the Results}
Throughout this comprehensive evaluation, it has been established that \ac{PLITA} successfully captures both invariant and tempo-variant features within a unified representation. Moreover, this capability allows \ac{PLITA} to achieve markedly enhanced results in a variety of downstream tasks, as detailed in Table \ref{tab:eval}. These findings provide robust evidence in favor of the hypotheses posited by this study:
(i) Tempo-variant features hold valuable information that is distinct from that of invariant features,
(ii) Merely utilizing tempo-variant features as a source of natural variability restricts the representation’s effectiveness in numerous downstream tasks,
(iii) These features can be integrated into the representations through the proposed $\mathcal{L}_{tv}$
 loss function, and
(iv) The inclusion of these features leads to a significant improvement in model performance in scenarios where tempo-variant features are crucial.

\section{Ablation and Sensitivity Studies}\label{sec:abla}

We have studied both the effect of incorporating the novel $\mathcal{L}_{tv}$ loss function as well as the role of the hyperparameters when computing it. 
Figure \ref{fig:effect} demonstrates that the incorporation of $\mathcal{L}_{tv}$ leads to a significant and positive impact on the tempo-variant related tasks. \\

\noindent The impact of the hyperparameters introduced, i.e the number of inputs from each recording (\textit{N}) and the window size (\textit{W}) have also been evaluated. Table \ref{tab:abla1} indicates that the selected configuration, highlighted in bold, achieves the best performance, but also that all configurations yield superior results compared with existing methods.

\begin{table}[h]
\centering
\resizebox{\columnwidth}{!}{%
\begin{tabular}{lc|lcccc}
\toprule
\rowcolor{gray!20}
\multicolumn{2}{c}{{\textbf{Downstream Task}}} & 
\multicolumn{2}{c}{{\textbf{AFIB Classification}}} & 
\multicolumn{2}{c}{{\textbf{Sleep Stage Classification}}} \\
 
\cmidrule(lr){1-2} \cmidrule(lr){3-4} \cmidrule(lr){5-6}

\multicolumn{2}{c}{\textbf{Pre-Train Dataset}} & 
\multicolumn{1}{c}{\textbf{SHHS}} & 
\multicolumn{1}{c}{\textbf{Icentia}} & 
\multicolumn{1}{c}{\textbf{SHHS}} & 
\multicolumn{1}{c}{\textbf{Icentia}} \\
 
\cmidrule(lr){1-2} \cmidrule(lr){3-3} \cmidrule(lr){4-4}
\cmidrule(lr){5-5} \cmidrule(lr){6-6}
\textbf{N} & \textbf{W (Seconds)} & 
\textit{Accu. (\%)} & 
\textit{Accu. (\%)} & 
\textit{Accu. (\%)} & 
\textit{Accu. (\%)}\\
    
\midrule
\rowcolor{gray!10}
3 & 120 & 79.5 & 77.6 & 74.3 & 73.7 \\

\textbf{4} & \textbf{120} & \textbf{80.7} & \textbf{80.0} & \textbf{75.3} & \textbf{74.8}\\

\rowcolor{gray!10}
5 & 120 & 77.3 & 78.6 & 73.2 & 73.9\\

4 & 90 & 76.6 & 77.8 & 74.1 & 74.5\\

\rowcolor{gray!10}
4 & 150 & 76.9 & 79.4 & 73.3 & 72.9\\
 
\bottomrule
\end{tabular}}
\vspace{5pt}
\caption{Sensitivity Study.}
\label{tab:abla1}
\end{table}

\section{Conclusions}
This research provides strong evidence that merely using the tempo-variant attributes of \ac{ECG} signals as a source of natural variability is insufficient. To overcome this, we introduce \ac{PLITA}, a novel \ac{SSL} technique for ECG analysis. By incorporating the $\mathcal{L}_{tv}$ loss function into the training objective, the model is directed to efficiently encode these tempo-variant attributes. This significantly enhances the model’s ability to excel in various tasks where such attributes are crucial. 

\paragraph{Limitations: }\ac{PLITA} has only been evaluated on a single architecture (\ac{ViT}). In addition, only BYOL has been utilized to capture the invariant features, leaving aside other non-contrastive frameworks such as \ac{DINO} \citep{dino} or \ac{VIC-REG} \citep{vicreg}. However, the incorporation of $\mathcal{L}_{tv}$, is agnostic to any of these two we can hypothesize that a similar performance improvement will be obtained for any combination.

\paragraph{Broader Impact: } We believe that the incorporation tempo-variant information as a training objective will inspire not only future \ac{ECG} but also in general time series \ac{SSL} methods.


\bibliography{aaai25}


\end{document}